# Focused Specific Objects NeRF

Yuesong Li[1,*]   Feng Pan[1]   Helong Yan[2]   Xiuli Xin[1]   Xiaoxue Feng[1]

**Abstract**

Most NeRF-based models are designed for learning the entire scene, and complex scenes can lead to longer learning times and poorer rendering effects. This paper utilizes scene semantic priors to make improvements in fast training, allowing the network to focus on the specific targets and not be affected by complex backgrounds. The training speed can be increased by 7.78 times with better rendering effect, and small to medium sized targets can be rendered faster. In addition, this improvement applies to all NeRF-based models. Considering the inherent multi-view consistency and smoothness of NeRF, this paper also studies weak supervision by sparsely sampling negative ray samples. With this method, training can be further accelerated and rendering quality can be maintained. Finally, this paper extends pixel semantic and color rendering formulas and proposes a new scene editing technique that can achieve unique displays of the specific semantic targets or masking them in rendering. To address the problem of unsupervised regions incorrect inferences in the scene, we also designed a self-supervised loop that combines morphological operations and clustering.

## 1、Introduction

Nowadays, an increasing number of tasks require understanding and interaction with scenes, posing higher demands for scene reconstruction and novel view synthesis, which not only involves restoring the geometric structure of a scene but also identifying its semantic information. Hence, NeRF with semantics has been a hot research topic. This paper makes important improvements in training speed and scene editing based on the ideas of Semantic-NeRF[2].

In the practical application of NeRF, we have found that its rendering effect is often not ideal due to multiple reasons, among which input ray order and data quality are the most fatal. In order to improve the universality of NeRF, this paper analyzes the important factors that affect rendering performance in Section 4. On the one hand, the order of input rays in training must be randomized and cannot be input according to the corresponding pixel arrangement order in the image, otherwise it will greatly reduce the rendering quality.

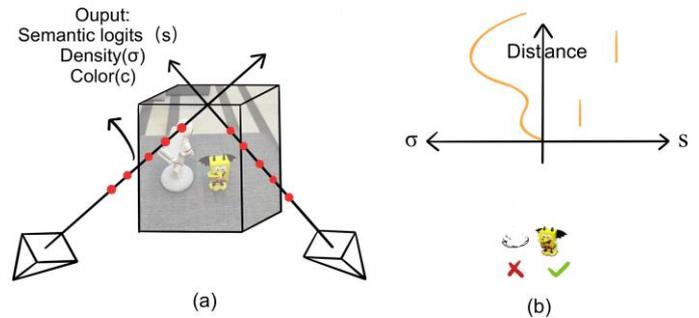

**Fig.1** An Overview of the scene editing method proposed in this paper: (a) demonstrates the emitted rays and their sampling points. Top (b) is the probability density function of the volume density $\sigma$ and the maximum semantic probability over the ray travel distance. Bottom (b) is the role of the self-supervision loop

On the other hand, the scene semantic prior is very helpful for NeRF to reconstruct geometric structures, and can even independently restore the scene structures without pixel colors. This can make up for the limitations of learning solely from pixel colors, such as low light, high light, and blurry images in the trainsets that can affect the reconstruction of scene structures.

[1]Beijing Institute of Technology, Beijing, China.

[2]The 28th Institute of China Electronics Technology Group Corp, Nanjing, China.

[*]Corresponding author. E-mail: dcch_lys@163.com

Many application scenarios such as product displays and portrait photography only require NeRF to reconstruct the structure of a single or partial object, followed by downstream tasks such as novel view synthesis and 3D modeling. However, most NeRF models focus on full scene learning. And if the background is complex and there are too many objects in the scene, it can easily lead to longer learning time and poorer rendering performance. Therefore, a new technique is proposed in section 5.3 to focus the network on the specific targets and train quickly without being affected by complex backgrounds. This technique divides all rays into positive and negative samples according to semantic labels. The rays corresponding to a certain class or several types of semantic labels are set as positive sample classes, while the rays with the other semantics are set as negative sample classes. Then, the colors of all negative sample ray are reset to a color with high contrast to the average color of all positive sample rays. This unique resetting operation greatly accelerates the training speed and avoids the avoids the difficulty or even failure of learning the target object when the two colors are too similar. In the end, the training time on the self-built dataset[Online Resource 1] was reduced by 7.78 times using this method, and the smaller the objects, the greater the speed improvement. In addition, the rendering quality, measured by both PSNR and SSIM, was significantly better than that of Semantic-NeRF and [1], with an SSIM improvement of about 0.15.

This paper also extends the pixel semantic and color rendering formulas by adding a mask to filter the sampling points on the ray. The mask sets the volume density $\sigma$ of the sampling points whose maximum semantic probability is certain labels to zero, which has two effects: unique displays of the specific semantic targets and masking them. In the technique of unique displays, the final label of a ray is obtained by integrating all the filtered sampling points on the ray according to their weights, and then only the colors of the rays with the final label as the specific semantic labels are rendered. This design effectively avoids the random noise of semantic labels on the ray, and the selective rendering method greatly reduces computational complexity and shortens rendering time.

To address the problem of incorrect inference for non-semantic labels areas that are occluded in the scene, we design a self-supervised loop that combines morphological operations and K-means++ clustering[41] at the output end of the Focused Specific Objects NeRF(abbreviated as FSO-NeRF) to correct network output. All presented in Section 5 are applicable to all NeRF-based models.

Previous research on 2D/3D segmentation has largely relied on dense semantic labeling of datasets, which is so heavy that the cost is unacceptable. Therefore, this paper also studies the weak supervision. Considering the inherent multi-view consistency and smoothness of NeRF, it has good semantic and color generalization capabilities. In some simple scenes, semantic labels can be extended to the whole scene with only a few clicks. This is also validated in Section 6.4, Figure 22, where even with sparse semantic labels, FSO-NeRF renders better. Additionally, as shown in Section 6.4, Figure 23, it is found that using negative sample rays with a sparse sampling rate of around 0.15 in fast training technique actually produces the best rendering quality.

2、**Related Work**

FSO-NeRF is influenced by previous studies on implicit representations of 3D geometry and 3D semantic segmentation. Below is a brief review of the relevant work in these fields.

## 2.1 Semantic Segmentation

Computer vision aims to achieve high-level semantic understanding of 3D scenes captured by digital images and videos. This is fundamental for applications such as visual navigation[6] and robotic interaction[5]. However, the most common approach[4, 8, 9, 11, 17, 18] to scene understanding focuses on 2D reasoning and only produces per-pixel annotations, largely ignoring the underlying 3D structure of the scene. For example, DeepLab[8] trains a CNN to segment each pixel in an image.

Different methods exist for representing shape in 3D, such as RGB images[7, 10, 14, 16, 19, 20], point clouds[21, 22, 23, 24, 25], and dense voxel grids[12, 13, 15, 26, 27]. Unlike these approaches, our method doesn't need any ground truth 3D annotations or input geometry. Instead, we reconstruct and segment a dense 3D representation from 2D inputs and supervision alone.

## 2.2 Learned Implicit Representations

Earlier research focused on using occupancy fields [28] or signed distance functions [29] to predict the geometry of shapes. These models were trained using ground-truth geometry, such as by sampling points from the meshes in the ShapeNet dataset[30]. More recently, Mildenhall et al. proposed Neural Radiance Fields (NeRF) [1], which predicts the density and colors of points in a given scene. This enables differentiable rendering by tracing rays through the scene and integrating over them, with supervision needed from only a small number of posed images. Estimating semantic labels of a 3D scene is closely related to predicting its geometry and appearance. Estimating the semantic labels of a 3D scene is closely related to predicting its geometry and appearance, the subsequent success of many follow-up works on NeRF [33, 34, 31, 32] has a long-lasting impact on semantic scene understanding. For instance, Semantic-NeRF [2] extends radiance fields to encode both semantics and geometry, allowing for smoother and more consistent semantic prediction through its multi-task learning setting. In some studies [2, 35, 36, 37], a scene was separated into a set of NeRFs associated with foreground objects, distinct from the background. But these methods[2, 35] cannot produce the 3D segmentation of objects necessary for rendering and editing individual objects.

## 3、Method

### 3.1 Architecture

FSO-NeRF is similar to Semantic-NeRF[2] in network structure. Given multiple images of a static scene with known camera intrinsics and extrinsics, FSO-NeRF uses MLPs to implicitly represent the continuous 3D scene density $\sigma$, color $c$, semantic label logits $s$. Input is continuous 5D vectors of spatial coordinates $(\mathbf{x}, \mathbf{y}, \mathbf{z})$ and viewing directions $\mathbf{d} = (\boldsymbol{\theta}, \boldsymbol{\varphi})$ after positional encoding. Semantic label logits $s$ can then be transformed into multi-class probabilities through a softmax normalisation layer.

Specifically, $\sigma = F_1(x), s = F_2(x)$ is designed to be an inherently view-invariant function of only 3D position, while the $c = F_\Theta(x, d)$ is a function of both 3D position and viewing direction. where $F_\Theta, F_1, F_2$ represents the learned MLPs.

To compute the color of a single pixel, FSO-NeRF approximates volume rendering by numerical quadrature with hierarchical stratified sampling. Within one hierarchy, if $r(t) = o + td$ is the ray emitted from the center of projection of camera space through a given pixel, traversing between near and far bounds ($t_n$ and $t_f$), then for selected K random quadrature points $\{t_k\}_{k=1}^{K}$ between $t_n$ and $t_f$. The approximated expected color is given by:

$$C(r) = \sum_{k=1}^{K} \hat{T}(t_k)\alpha(\sigma(t_k)\delta_k)c(t_k) \quad (1)$$

$$S(r) = \sum_{k=1}^{K} \hat{T}(t_k)\alpha(\sigma(t_k)\delta_k)s(t_k) \quad (2)$$

$$\hat{T}(t_k) = \exp\left(-\sum_{k'=1}^{k-1} \sigma(t_k)\delta_k\right) \quad (3)$$

where $\alpha(x) = 1 - \exp(-x)$, and $\delta_k = t_{k+1} - t_k$ is the distance between two adjacent quadrature sampling points.

## 3.2 Network Training

We train the whole network from scratch under photometric loss $L_P$ and semantic loss $L_S$:

$$L_p = \sum_{r \in R}\left[\left\|C_c(\mathrm{r}) - C(r)\right\|_2^2 + \left\|C_f(r) - C(r)\right\|_2^2\right] \quad (4)$$

$$L_s = -\sum_{r \in R}\left[\sum_{l=1}^{L} y_l(1 - \hat{p}_c^l(r))^\gamma \log \hat{p}_c^l(r) + \sum_{l=1}^{L} y_l(1 - \hat{p}_f^l(r))^\gamma \log \hat{p}_f^l(r)\right] \quad (5)$$

Where R are the sampled rays within a training batch, and $C(r)$, $C_c(r)$ and $\hat{C}_f(r)$ are the ground truth, coarse volume predicted and fine volume predicted RGB colors for ray $r$, respectively. Similarly, $\hat{p}_c^l$ and $\hat{p}_f^l$ are coarse volume and fine volume predictions for ray $r$, where L are all semantic label categories. Considering the imbalance of semantic label categories(Fig 4), focal loss is chosen as $L_s$ to encourage the rendered semantic labels to be consistent with the provided labels, whether these are ground-truth, noisy or partial observations. Additionally, if the ground truth label of ray r is equal to $l$, then setting $y_l$ to 1, otherwise $y_l$ to 0. Focal loss is not sensitive to $\gamma$ which is taken as 1 in the experiment. The total training loss $L$ is:

$$L = L_p + \lambda L_s \quad (6)$$

where $\lambda$ is the weight of the semantic loss and is set to 0.04 to balance the magnitude of both losses. When testing, we have discovered that the actual performance is not significantly influenced by the value of $\lambda$. By utilizing photometric and semantic losses, the network is naturally incentivized to produce consistent 2D renderings across multiple views.

## 4、Factors Affecting Rendering Performance
### 4.1 Input Ray Order

For two common order of light input to be trained, either following the order of the image rows and columns or a random order, the former has a strong negative impact on rendering at high resolution, leading the network to a bad local optimum, as shown in Figure 2,3 below.

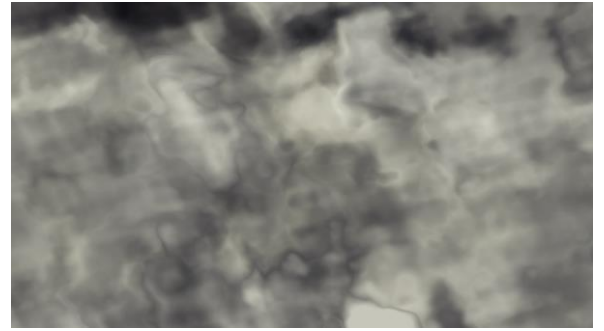

**Fig.2** Bad Rendering Effect: When the light is arranged according to the rows and columns in the picture, the rendering results of the FSO-NeRF on the testsets after training for 200,000 iterations

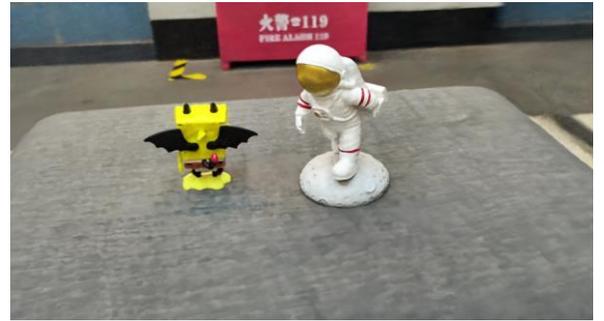

**Fig.3** Good Rendering Effect: When the rays are randomly arranged, the rendering results of the FSO-NeRF on the testsets after training for 200,000 iterations

A reasonable explanation for this phenomenon is that the network's learning rate is large in the first tens of thousands of rounds, making it the most important part of the entire training process. Random input order allows the network to have a comprehensive perception of the overall scene first and then refine local details in later rounds, even pixels without supervised labels can be estimated by the NeRF's generalization capabilities. However, inputting rays according to the row and column order of the image will make the network to focus on learning the background rays which account for a large proportion in the initial training rounds. This is detrimental to learning the overall scene, as object-corresponding rays account for a small proportion of the total, as shown in Figure 4. In the self-built dataset proposed in this paper, the object-corresponding rays account for about 2% of the total rays in one image on average, as indicated by the red line. Additionally, the training image size is $960 \times 540$, the trainsets have a total of 89 images, and each round of input ray batch is 1024, requiring 45056 rounds to iterate through the entire trainsets.

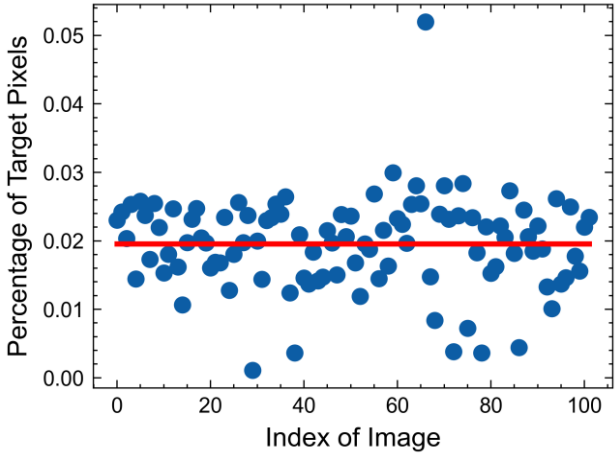

**Fig.4** Percentage of Target Pixels: In the self-made dataset, the proportion of non-background pixels in all pixels of a single image

**4.2 Scene Semantic Priors**

Another important factor is the high correlation between the scene semantic priors and the geometric structure of the scene, and adjacent identical semantic labels often come from the same object, so the semantic priors can help NeRF learn the volume density $\sigma$ of the scene.

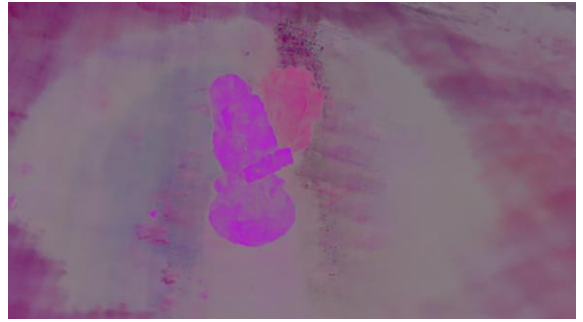

**Fig.5** Scene Structure: It is recovered by FSO-NeRF without pixel colors but pixel semantic labels only

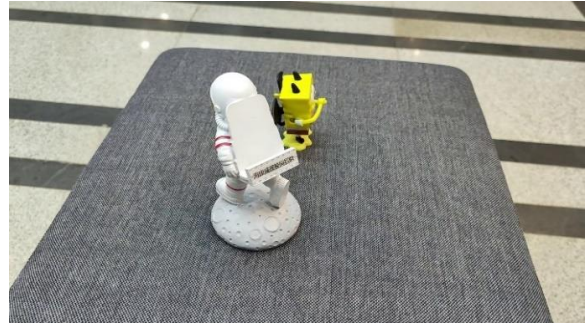

**Fig.6** The Ground Truth of Fig.5

Our method effectively utilizes this property by using both color and semantic priors to avoid the blurriness or artifacts that may appear when reconstructing a scene solely based on colors, as shown in Figure 7. This is particularly common in scenes with significant highlights and shadows.

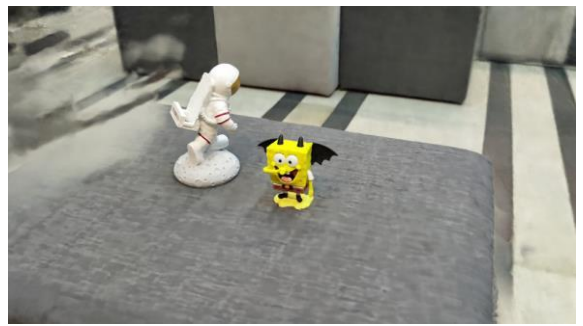

**Fig.7** Artifact in Upper Left

In the initial training phase, we only calculate the semantic loss of rays in the scene without the color loss to quickly rebuild the general scene structure. Then, the semantic and color loss of rays in the scene is calculated, to finetune FSO-NeRF with more detailed color information, in order to refine and perfect the structure of the scene.

## 5、Scene Editing Technique

The scene editing technique proposed in this paper utilizes the predicted semantic information of the scene and includes two rendering effects: unique displays of the specific semantic targets or masking them in rendering(Often used to remove occlusion).

### 5.1 Unique Displays

We first generalized the pixel semantic and color rendering formulas by adding masks to equation 8 and 9, where '*ob*' is the category number of the specific semantic targets. The mask sets the volume density $\sigma$ of the sampling points with the max semantic probability of non-'*ob*' or '*ob*' to zero.

$$S(\mathrm{r}) = \sum_{k=1}^{K} \hat{T}(t_k) \alpha\left(\sigma(t_k)\delta_k\right) s(t_k) M_k \quad (7)$$

$$C(\mathrm{r}) = \sum_{k=1}^{K} \hat{T}(t_k) \alpha\left(\sigma(t_k)\delta_k\right) c(t_k) M_k \quad (8)$$

$$M_k = 1 - \left|\mathrm{sign}(ob - \mathrm{s}(t_k))\right| \quad (9)$$

However, it is observed that there are some semantic noise points with random semantic labels on any ray corresponding to a pixel. These semantic noise points affect the rendering effect after applying formulas 8, 9, and 10. As a result, the rays with non-specific semantic label '*ob*' will also render colors after integrating according to probability density parameters.

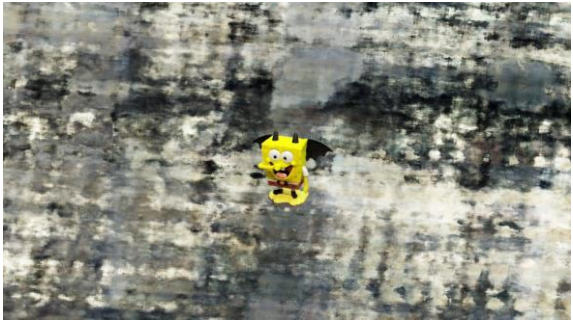

**Fig.8** Rendering Result Affected by Semantic Noise

To avoid the influence of semantic noise points, we improved the algorithm as shown in Algorithm 1. First, we only keep the volume density of the sampling points with the maximum semantic probability as the specific semantic '*ob*' and background is preserved, and the rest of the sampling points are set to zero. Then we calculate the maximum semantic probability category for all rays after integration. If the category equals the specific semantic '*ob*', we render the color of the corresponding pixel for that ray. Considering that only a fraction of pixels in an image have the specific semantic label of '*ob*', this selective rendering method can significantly reduce computational complexity and shorten rendering time, especially for small to medium sized targets in images.

**Algorithm 1:** The improved unique displays of the specific semantic targets

Inputs:
Let the prediction of NeRF is raw.
Let the semantic probability of raw is S.
Let the classes of the specific semantic targets and background are respectively '*ob*', '*bg*'.
Let the volume density of points in rays is $\sigma$.
The function F(raw) can convert the raw to the integrated semantic logits of rays.
The function G(raw) can convert the raw to the integrated colors of rays.
**Results:**
The colors of the specific semantic targets.
All semantic labels in the scene.

**1** points_labels ← argmax(softmax(S))
**2** mask ← zeros_like(points_labels)
**3 for** point **in** all rays **do**
**4**   **if** points_labels[point] ==*ob* or points_labels[point] == *bg*
**5**     **then**
**6**       mask[point] ← 1
**7**   **end**
**8 end**
**9** $\sigma \leftarrow \sigma$ * mask
**10** seman_labels ← argmax(softmax(F(raw)))
**11** color ← ones_like(seman_labels)
**12 for** ray **in** all rays **do**
**13**   **if** seman_labels[ray] == *ob*
**14**     **then**
**15**       color[ray] ← G(raw)
**16 end**

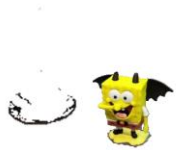

**Fig.9** The RGB Result of The Improved Algorithm

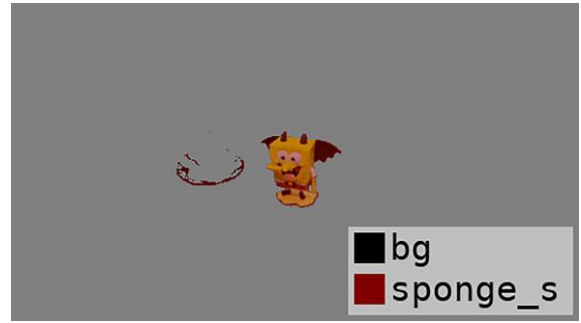

**Fig.10** The Semantic Result of The Improved Algorithm

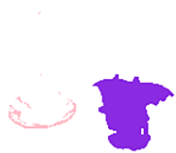

**Fig.11** The Result After Morphological Opening and Clustering

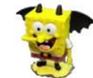

**Fig.12** The Result After Self-supervised Loop Correction

Figures 9 and 10 show the rendering results of the improved method. Compared with Figure 13, it is found that the FSO-NeRF has an error in inferring the semantic labels of the shielded astronaut's base. This phenomenon often occurs in unsupervised label regions caused by occlusions. To address this issue, a self-supervised method combining morphological operations and K-means++ clustering was designed. Due to the strong generalization reasoning capabilities of FSO-NeRF, even if there are incorrect inferences in some regions, this self-supervised method can correct the inferences. The corrected output is shown in Figure 12.

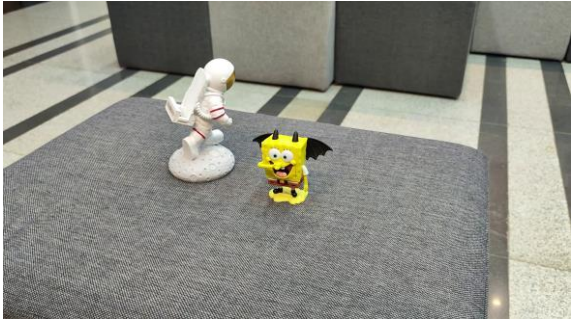

**Fig.13** The Ground Truth

During the training process, FSO-NeRF is first trained for a certain number of epochs to obtain an initial perception of the scene structure, semantics, and colors. In the self-made dataset tested in this paper, continuous observation of the network rendering output shows that the scene semantic information is basically formed after 30,000 epochs (ray batch:1024), and the scene color information is basically formed after 20,000 epochs. Therefore, the initial training epochs can be set to 30,000. After the initial training, at intervals of a certain number of epochs (2500 epochs in this paper), the specific semantic targets are uniquely displayed on the trainsets using the scene editing technology described in this section. Then, these newly rendered images are sequentially subjected to morphological opening and closing operations and clustering to correct the rendering results. Finally, the corrected results are used as semantic supervised labels for FSO-NeRF.

**5.2 Rendering with Mask**

To mask the specific semantic targets, all that is needed is to modify $M_k$ in Formula 10 to the following:

$$M_k = |\text{sign}(ob - \text{s}(t_k))| \quad (10)$$

where '$ob$' is the specific semantic label to be masked. The results can be seen in Section 6.3.

**5.3 Fast Training Technique**

So far, most NeRF models have been designed to learn the entire scene. However, When there are many objects in the scene and the background is complex, it often leads to longer learning time, poorer rendering performance, and even many artifacts, as shown in Figure 7 of Section 4.2.

To address this, we have designed a fast training technique for FSO-NeRF that focuses on the specific semantic targets without being affected by complex backgrounds, which greatly reduces training time. The shortened training time in this technique is related to the α indicator (Formula 11), where a smaller α value leads to faster training. This indicator reflects the average pixel area percentage of the specific targets in the image. For more detailed time comparisons, see Section 6.2. On our self-made dataset, this technique has increased the training speed by 7.78 times and is applicable to all NeRF-based models.

$$\alpha = \frac{\sum_{i=1}^{N} n_{ob}}{H * W * N} \quad (11)$$

Specifically, the fast training technique first discards rays that do not have semantic labels, and then divides the remaining rays into two classes based on their semantic labels. Rays with labels corresponding to the specific semantic targets are considered as positive samples, while rays with other labels are considered as negative samples. The semantic labels of both positive and negative samples are directly learned by the FSO-NeRF. After multiple experiments, we find that directly ignoring the colors of negative sample rays can result in poor rendering performance (Figure 14), as there are no color comparisons between positive and negative samples.

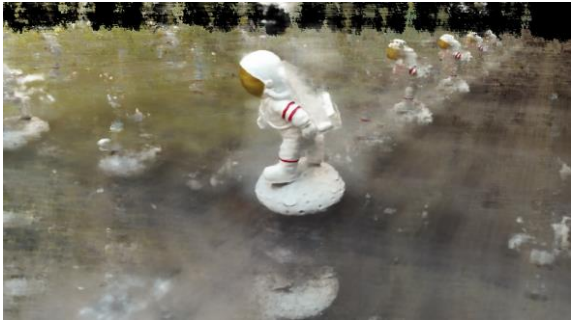

**Fig.14** The Rendering Result Without Color Contrast of Positive and Negative Sample

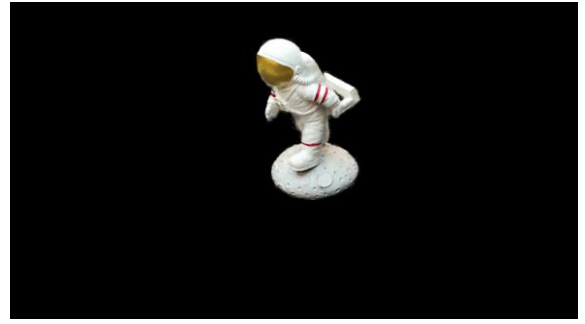

**Fig.15** The Rendering Result of The Improved Algorithm

Our goal is to perform 3D reconstruction and novel view synthesis of the specific semantic targets, and we do not care about colors other than the specific targets'. Therefore, we reset the colors of all negative sample rays. The larger the color differences between the reset colors and the colors of the specific semantic targets, the better the rendering effect (see Figure 15). When the two colors are similar, it is difficult or even impossible for the FSO-NeRF to learn the semantic targets. In addition, the color resetting operation reduces the number of labels to be learned, and speeds up the training process. Finally, starting from the average color $(R_1, G_1, B_1)$ of the specific semantic targets, we use a low computational cost approximation of color difference $\Delta C$ (formula 12) and a depth-first search in the CIELAB color space to determine an adaptive high color difference reset color $(R_2, G_2, B_2)$.

$$\Delta C = \sqrt{(2+\frac{\tau}{256}) \times (R_1 - R_2)^2 + 4 \times (G_1 - G_2)^2 + (2+\frac{255-\tau}{256}) \times (B_1 - B_2)^2} \quad (12)$$

$$\tau = \frac{R_1 + R_2}{2} \quad (13)$$

## 6、Experiments
### 6.1 Datasets

We have two datasets, both of which are spherical 360 scenes. One is from [1] paper, and the other is a self-made dataset[Online Resource 1]. The self-made dataset has a more complex background, with brightly colored fire hydrants and highlights reflected on the floor, as well as multiple occlusion relationships between the two objects. Under the self-made complex dataset, the experimental results can better demonstrate the success of FSO-NeRF improvement.

The experimental environment configuration of this paper is as follows: Intel Xeon E5-2670 V3 64GB, Tesla V100 16GB GPU, Ubuntu 20.04 operating system, and Pytorch deep learning framework.

### 6.2 Fast Training Technique

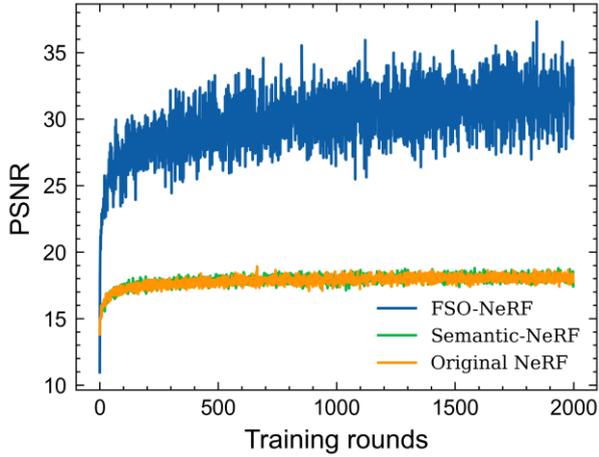

**Fig.16** The PSNR Curve: Calculated every 100 iterations during the training

The Fast training legend is the fast training technique described in section 5.3. From Figure 17, it can be seen that the PSNR calculated on the ray batch during training is consistently better than that of Original NeRF [1] and Semantic-NeRF [2]. However, considering the limitations of PSNR metric that only measures the relationship between the maximum signal and background noise, without quantifying the structural similarity between images, we add the SSIM metric for comparison. We provide detailed explanation of the SSIM metric in Supplementary Information section A.

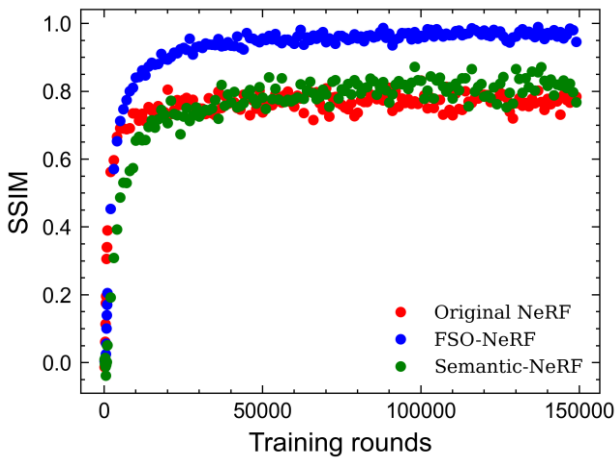

**Fig.17** The SSIM in Training Rounds: The vertical axis is the average SSIM in the testsets which rendered by FSO-NeRF after trained for the training rounds displayed on the horizontal axis

From Figure 18, the SSIM metric of the fast training technique is approximately 0.15 higher than other methods at 150,000 iterations and only requires 15,000 iterations to achieve the same SSIM value of Semantic-NeRF at 150,000 iterations. Combined with Table 1, under the same rendering effect, the methods based on fast training technique can even improve the training speed by 7.78 times compared to the original model(Semantic-NeRF), greatly shortening the training time. We also conducted experiments based on Mip-NeRF.

**Tabel.1** Training Time of Methods

| *SSIM=0.8* | *Time(hour)* |
| --- | --- |
| FSO-NeRF (based [1]) | 1.23h |
| Semantic-NeRF | 9.57h |
| Original NeRF | 8h |
| FSO-NeRF (based [38]) | **1.12h** |
| Mip-NeRF | 6.88h |

Finally, the fonts in the scene was selected as the rendering details for comparison. Clearly rendering these fonts is technically difficult, but our fast training technique reproduces these fonts very well, with results far superior to Original NeRF and Semantic-NeRF.

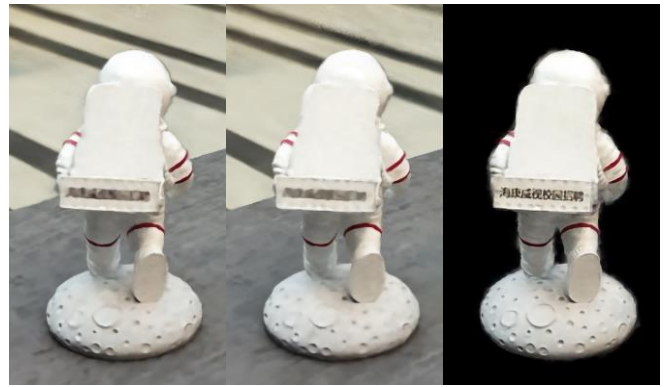

**Fig.18** Comparison of Rendering Results in 200000 Training Rounds: From left to right are Semantic-NeRF, Original NeRF, FSO-NeRF

### 6.3 Scene Editing Technique

Sections 5.1 and 5.2 introduce the principles of two rendering effects in scene editing technology: unique displays of the specific semantic targets or masking them in rendering. The unique displays are demonstrated in section 5.1. This section supplements the display effect of masking the specific semantic targets (Figure 20, 21).

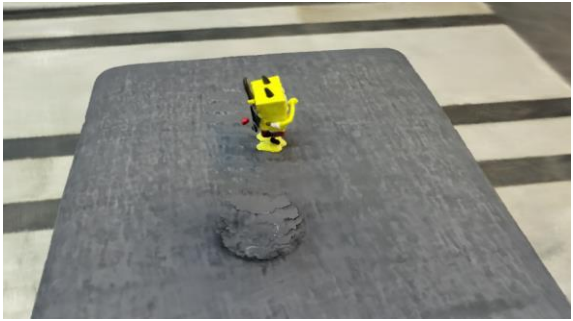

**Fig.19** The Result of Blocking a Specific Semantic Target

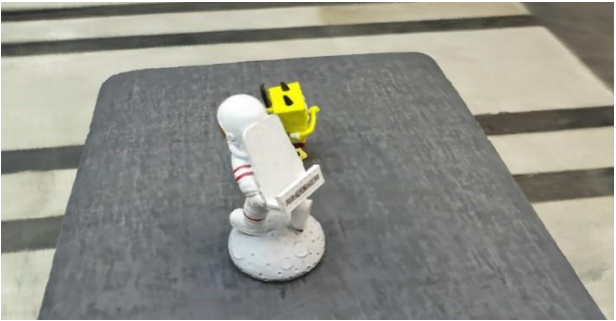

**Fig.20** The Comparison

Although the structure and color restoration of the occluded SpongeBob is good, the structure below the astronaut base is not smooth enough.

### 6.4 The Improvement of Weakly Supervised Learning

The workload of annotating dense semantic labels for datasets is significant. This paper also researches and conducts experiments on weakly supervised learning. Considering the inherent multi-view consistency and smoothness of FSO-NeRF, the semantics and colors in the scene can be well generalized to areas without supervised labels, which greatly saves the workload. Even in simple scenes, it only takes a few rough labeled points to extend the semantic labels to the entire scene.

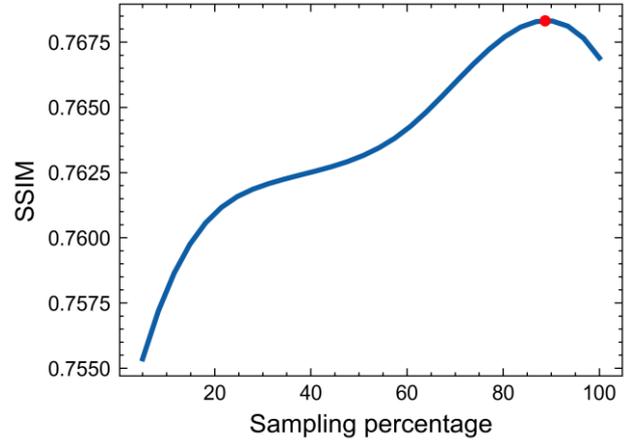

**Fig.21** The SSIM of Different Sampling Percentage: The horizontal axis is the proportion of the sampled semantic labels to the total semantic labels, and the vertical axis is the average SSIM in the testsets which rendered by FSO-NeRF after trained for 100,000 rounds

Figure 22 illustrates that sparse semantic labels and pixel colors can effectively reconstruct the scene structure and recover the corresponding semantic and color information. Even as shown by the red dot in the figure, when 88% sampling of the semantic labels as supervision, the rendering quality of the testsets is better than using all the semantic labels as supervision. The reasonable explanation is that moderately sparse labels introduce noise, which limits the flow of information in the network, prevents overfitting of the network, and leads to better rendering quality.

In addition, the fast training technique in Section 5.3 can further improve the rendering quality or reduce the training time under the same rendering quality through weakly supervised learning. Specifically, by sparsely sampling negative class rays and using them to train the FSO-NeRF, the SSIM metric on the testsets can be calculated as shown in Figure 23 and 24. It can be observed that when too many sampled negative class rays with reset colors(sampling ratio above 40%), this will seriously impede network learning. Therefore, moderately sampling negative class rays yields the best rendering effect. In this paper, the recommended sampling rate is 0.15.

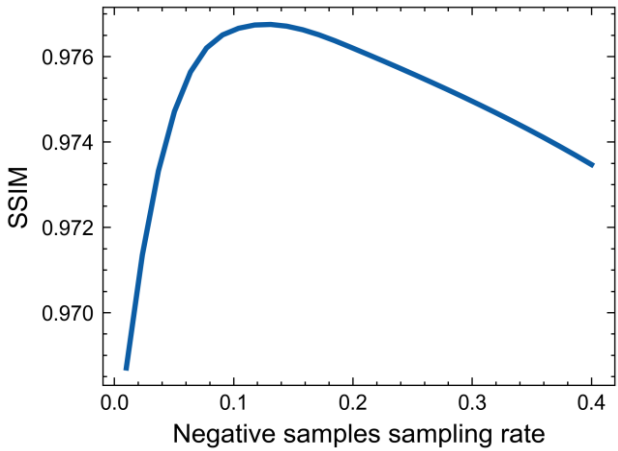

**Fig.22** The SSIM of Different Negative Samples Sampling Rate: The vertical axis is the average SSIM in the testsets which rendered by FSO-NeRF after trained for 100,000 training rounds and the horizontal axis is the sampling rate of negative rays

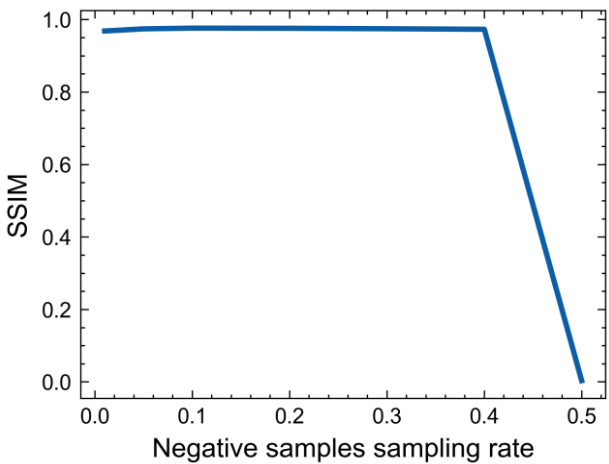

**Fig.23** The SSIM of Different Negative Samples Sampling Rate: The vertical axis is the same as the Fig.23 above, but the horizontal axis has a range of 0 to 0.5

**7、Conclusion and Future Work**

This paper presents a new scene editing technique that extends the pixel semantic and color rendering formulas. The technique achieves the unique displays or masking the specific semantic targets. In addition, to address the problem of inference errors in some unsupervised region, a self-supervised loop was designed at the output end of the FSO-NeRF. Our experiments demonstrate the effectiveness of the scene editing technique.

By focusing the FSO-NeRF on the specific semantic targets, the training speed was increased by 7.78 times, overcoming issues such as longer learning times and poorer rendering quality in complex scenes. Finally, weakly supervised learning is also used to reduce the amount of annotation work.

An interesting direction for future research is unsupervised 3D semantic segmentation based on NeRF which takes advantage of the geometric priors of LLM.

**Acknowledgements**

In paper writing, I once had unsatisfactory results and felt quite frustrated. I am grateful for the support and encouragement from my family during that time, as well as the guidance from my mentor Feng Pan and my classmates. Without any funding organizations, all funding is self-raised.

**Supplementary Information**

**A. An Introduction to Evaluation Metrics**

Structural similarity index (SSIM) is a metric used to quantify the structural similarity between two images. SSIM measures structural similarity in imitation of human visual system (HVS) and is sensitive to changes in local structure of images. SSIM quantifies image attributes from brightness, contrast, and structure, using mean to estimate brightness, variance to estimate contrast, and covariance to estimate structural similarity. The range of SSIM values is 0 to 1, with larger values indicating more similar images. If two images are exactly the same, the SSIM value is 1.

Given two images of x and y, the illumination, contrast, and structure between them are shown in the following formulas, where $\mu_x$ is the average value of x, $\sigma_x^2$ is the variance of x, $\mu_y$ is the average value of y, $\sigma_y^2$ is the variance of y, $\sigma_{xy}$ is the covariance of x and y. $c_1 = (k_1 L)^2$ and $c_2 = (k_2 L)^2$ are two constants used to maintain stability avoiding zero division. L is the max value of the pixel colors (uint8 type is 255, floating point type is 1). In this paper, $k_1$=0.01, $k_2$=0.03:

$$l(x, y) = \frac{2\mu_x \mu_y + c_1}{\mu_x^2 + \mu_y^2 + c_1} \quad (14)$$

$$c(x, y) = \frac{2\sigma_x \sigma_y + c_2}{\sigma_x^2 + \sigma_y^2 + c_2} \quad (15)$$

$$s(x, y) = \frac{\sigma_{xy} + c_3}{\sigma_x \sigma_y + c_3} \quad (16)$$

$$SSIM(x, y) = [l(x, y)^\alpha \cdot c(x, y)^\beta \cdot s(x, y)^\gamma] \quad (17)$$

Peak Signal to Noise Ratio (PSNR) is a measurement standard for evaluating image quality, but it has limitations: it only measures the reference between the maximum signal and background noise, without considering the visual recognition and perception characteristics of the human eye. Therefore, the evaluation results of PSNR often differ from subjective human perception. The unit of PSNR is dB, with a higher value indicating less image distortion. Generally speaking, a PSNR above 40dB indicates that the image quality is almost as good as the original image; Between 30-40dB, it usually indicates that the distortion loss of image quality is within an acceptable range; Between 20-30dB indicates poor image quality; PSNR below 20dB indicates severe image distortion.

Given a groundtruth grayscale image I and a predicted grayscale image of size m×n. The PSNR formula is as follows, where $MAX_I$ is the max value of the pixel colors (uint8 type is 255, floating point type is 1):

$$MSE = \frac{1}{mn} \sum_{i=0}^{m-1} \sum_{j=0}^{n-1} [I(i,j) - K(i,j)]^2 \quad (18)$$

$$PSNR = 10 \cdot \log_{10}(\frac{MAX_I^2}{MSE}) \quad (19)$$

For the PSNR of RGB images, this paper first calculates the mean square deviation on all RGB channels, and then uniformly calculates the PSNR.

**B. Datasets**

The self-built dataset has also been uploaded as supplementary information[Online Resource 1], detailed in Section 6.1 Datasets.

**Additional Results**

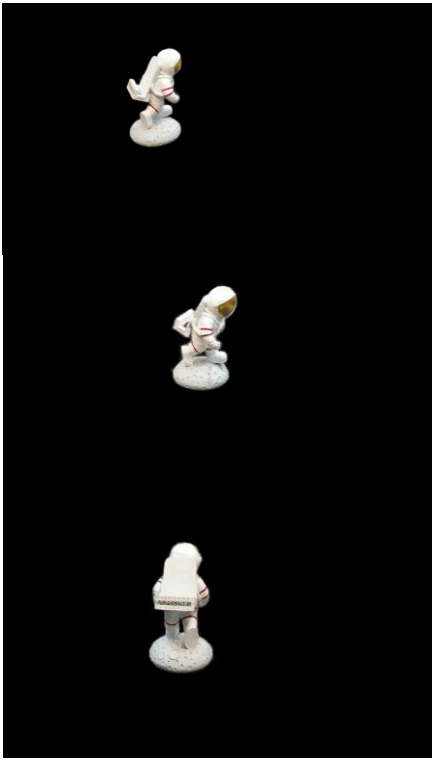
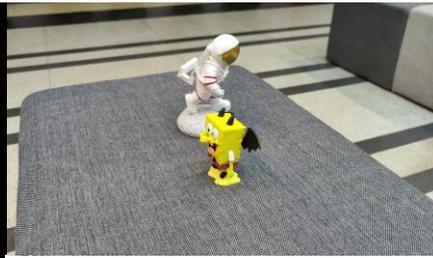
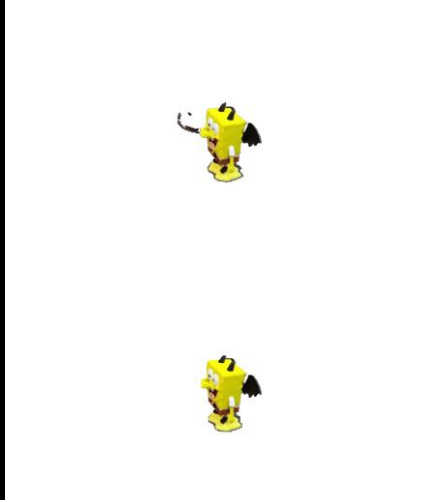
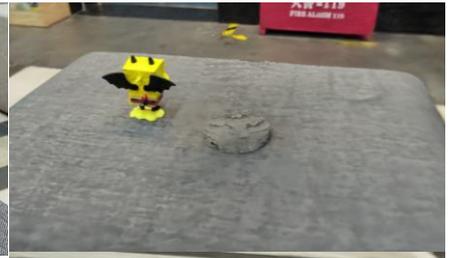
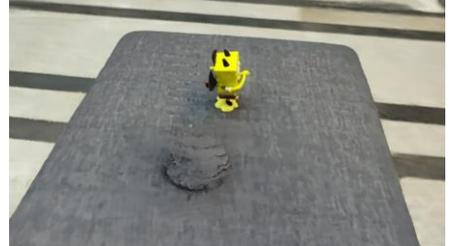
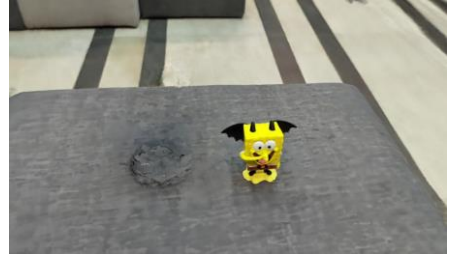

**Fig.24 25 26** More Results of the Fast Training Technique

**Fig.27 28 29** More Results of Unique Displays

**Fig.30 31 32** More Results of Masking the Specific Targets